# YouTube-8M Video Understanding Challenge Approach and Applications

Edward Chen
The Pennsylvania State University
University Park, Pennsylvania 16802
edwardjchen@gmail.com

## Abstract

*This paper introduces the YouTube-8M Video Understanding Challenge hosted as a Kaggle competition and also describes my approach to experimenting with various models. For each of my experiments, I provide the score result as well as possible improvements to be made. Towards the end of the paper, I discuss the various ensemble learning techniques that I applied on the dataset which significantly boosted my overall competition score. At last, I discuss the exciting future of video understanding research and also the many applications that such research could significantly improve.*

## 1. Introduction

To aid with the advancement of machine learning and computer vision research, large and varied datasets are necessary for effective training results. With the creation of image datasets such as the Caltech 101 [4], PASCAL [5], and ImageNet [6], image understanding research has greatly accelerated and is at a point far beyond what it would have been without the existence of such datasets. In addition to image understanding, there has also been a major shift of focus towards video understanding research. Datasets such as Sports-1M [7] and ActivityNet [8] have paved the way for providing large video benchmarks but are limited to solely activity and action categories - with about 500 categories total [8]. In an effort to further the advancement of video benchmarks, Google Research released the YouTube-8M [9] dataset with extensive features such as:

- an increase in the number of annotation classes - 4800 Knowledge Graph entities [9]
- a substantial jump in the amount of labeled videos - over 8 million videos [9]
- a large-scale video annotation and representation learning benchmark [9]

In addition to such characteristics, Google Research has also provided pre-computed audio-visual features for the 1.9 billion video frames - which are meant to significantly level the playing field for all levels of researchers.

To accelerate research and understanding on the YouTube-8M dataset, Google Research hosted a large-scale video classification challenge as a competition on Kaggle.com with $100,000 in prize money. In addition to being sponsored by Google Cloud, which provided competitors with $300 in Google Cloud credits, Google Research also released open-sourced starter code based on TensorFlow. With providing such resources, the goal of Google Research is to accelerate research on large-scale video understanding, noisy data modeling, and to further the understanding of various modeling approaches and their strengths and weaknesses in such a domain [9].

As a competitor in the Kaggle competition, I was fortunate to receive such resources to test various modeling approaches to find the highest performing single or set of models. In the rest of this paper, Section 2 will describe the performance metrics and development environment used for the competition. Section 3 will describe each of the notable models that I tested on the dataset. Section 4 will detail the approaches I used for fusion modeling towards the end of the competition, and Section 5 will cover possible applications of the YouTube-8M dataset and Kaggle competition results. Section 6 will detail any additional ideas I have. I offer concluding remarks with Section 7.

## 2. Performance metrics and development environment

The primary metric used for evaluating performance in this competition was the Global Average Precision (GAP) at k. The formula used to calculate the Global Average Precision is

$$GAP = \sum_{i=1}^{N} p(i) \Delta r(i)$$

where N is the count of final predictions, p(i) is the precision, and r(i) is the recall.

To obtain such a value, each competitor must submit a comma-separated values (CSV) file with 700,640 prediction rows and a header row. Each prediction row should contain a Video ID with a list of predicted labels and the corresponding confidence scores for each label.
A sample submission file is shown in Figure 1.

The primary language and technologies used for development were Python for scripting, PyCharm for developing, GitHub for hosting, TensorFlow for model-building, and the Google Cloud Platform for training, evaluating, and predicting. To get familiar with the Google Cloud Platform, I followed the "Getting Started With Google Cloud" tab on the Kaggle competition's home page.

## 3. Models

For the competition, participants were provided with two sets of the data - video-level and frame-level. The video-level data has a total size of 31 GB. For the video-level data set, each video contains the following:

- Video ID: Unique ID for the video
- Labels: The list of labels corresponding to the video
- Mean RGB: A float array of length 1024
- Mean Audio: A float array of length 128

For all of my experimentation with the data, I used the video-level data set.

### 3.1 Logistic regression

The logistic regression model was implemented using the TF-Slim library for TensorFlow. The model contains a single fully-connected layer with a sigmoid activation function and L2 weight regularization value of 1E-8. The logistic regression model was first trained on solely visual data and received a public GAP score of 0.70627. After combining the visual data with the audio data, the model scored a GAP of 0.75884. I made another attempt at improvement by modifying the L2 weight regularization to be an L1 weight regularization instead - doing so raised the GAP by a minute amount up to 0.75901. One of the last attempts with the logistic regression model involved decreasing the L1 penalty to 1E-10 from 1E-8. The GAP score, again, increased by a fraction of an amount up to 0.75911.

Overall, the logistic regression model resulted in producing very strong results despite its simplicity. The goal was to determine whether tuning some of the hyperparameters of the logistic regression model would increase the GAP by any significant amount. Based on the results, the largest score jump occurred after combining the audio data with the visual data - it resulted in a major

```
VideoId,LabelConfidencePairs
100000001,1 0.5 2 0.3 3 0.1 4 0.05 5 0.05
etc.
```

Figure 1: Submission file format

.05257 jump in GAP. The diminutive increases, .00017 and .0001, in GAP from the changes in the weight regularization may have simply been random fluctuations as the changes are not significant enough to come to any conclusions.

A possible future experiment may entail testing many more different values for the regularization techniques in order to find the optimal value.

### 3.2 Mixture of experts

The out-of-the-box mixture of experts model contains a fully connected layer for both the gate and expert activations as well as a softmax activation layer for the gates and a sigmoid activation layer for the experts. The mixture of experts model is very flexible for accommodating a variable amount of experts, which simply requires modifying a flag in the command. The mixture of experts model with 2 mixtures resulted in a GAP score of .78010. According to the competition paper, the performance increases by 0.5%-1% on all metrics as the number of mixtures increases from 1 to 2, and then from 2 to 4 [9]. Such a claim was further evidenced by the score of .78629 for the model with 4 mixtures - an increase of .00619 or .619%. The model with 3 mixtures also showed an increase from that with 2 mixtures, but only of .00383. The performance seemed to further increase as the complexity went up even further. For the model with 5 mixtures, the score rose to .79018 and then to .79096 with 6 mixtures. With 7 mixtures, the performance went up to .79244, which is a significant increase of .01234 from the model with 2 mixtures.

For all of the previous attempts on improving the mixture of experts model, the base learning rate parameter was set to a value of 0.01. To see if modifying the learning rate would have any effect on the performance, I decreased it significantly to 5E-4 and then also added the validation set as part of the training for the mixture of experts model with 8 and 9 mixtures. From a theoretical standpoint, adding the validation set to the model training should increase the performance slightly as it would then have more varied examples to learn from. Despite so, the model score resulted in roughly 0.76 after the training step. Due to lack of time, the poor training performance led to me not pursuing it further towards the inference stage.

In an experiment to test if increasing the complexity and number of layers of the mixture of experts model would further increase the performance, I added a hidden fully connected layer before the fully connected layer for the expert activations in the original model. The input to the added layer is the model input and it consisted of 2048 hidden neurons. I will refer to this model as MOE C later on in the paper. After training the complex mixture of experts model with 2 mixtures on just the training set, the GAP score seemed to hover around the value of 0.82. After going through the inference stage, though, the score dropped down significantly to 0.777. Based on my understanding, a possible reason for such a score difference could be the model overfitting. To try and correct such an issue, I attempted to train the same complex model with both the training and validation set. After the training stage, the score seemed to be even higher at a value of ~0.84 but then dramatically decreased to 0.77001 after going through the predictions from the model.

Based on said results, a slightly higher learning rate around the value of .01 seems to be ideal for increasing the performance of the mixture of experts model by simply increasing the number of mixtures. A possible future experiment would be to further increase the number of mixtures until the score starts to decrease. Another observation is that the low base learning rate of 5E-4 seemed to severely harm the performance of the mixture of experts model - dropping it by roughly 3%. Another interesting observation came from the dramatic score changes when increasing the complexity of the mixture of experts model by adding an intermediate hidden layer before the expert activations layer. Such an increase in complexity seemed to greatly harm the performance of the model in this experiment. A possible future test would be to experiment with the number of hidden neurons, layers, and mixtures on the complex mixture of experts model.

### 3.3 Multilayer perceptron

The first neural network-based model I experimented with was a multilayer perceptron model. It was constructed using the TF-Slim Tensorflow library. The first multilayer perceptron model I constructed contained an input layer, 2 hidden layers, and an output layer with a softmax activation function. Each of the 2 hidden layers were fully connected layers that consisted of 2000 hidden neurons and ReLu activation. The GAP score from that model came out to be 0.67187. As another experiment, I increased the number of hidden neurons from 2000 to 3000 and kept the rest of the structure the same. After running the model, it received a score of 0.65256 - surprisingly worse than the previous one.

In an attempt to dramatically impact the performance of the neural network, I decreased the number of hidden neurons down to 512 on the first hidden layer and 256 on the second hidden layer. The final activation function was also modified to a sigmoid function as opposed to softmax. As a result of the changes, the score increased to 0.77 after the training step.

Additional attempts at improving the performance were made by adding residual/skip connections [10] to the neural network. In an early attempt at residual multilayer perceptron models, I created a neural network with an input layer, 5 hidden layers, and an output layer with sigmoid activation. The 5 hidden layers had the following hidden neuron counts: 784, 512, 512, 512, 256. The model also consisted of 2 residual connections - one from the input layer to the third hidden layer and another from second hidden layer to the fourth hidden layer. The selection of the residual connections were arbitrary. Upon testing the model, the GAP score resulted in being 0.783 - a significant increase from the previous models. To further test the performance of residual connections, I created an additional model that was much deeper - 1 input layer, 9 hidden layers, and 1 output layer with sigmoid activation. The first hidden layer consisted of 1536 hidden neurons while the rest all contained 1024 hidden neurons. Residual connections were made between the following layer pairs: (0, 3), (2, 4), (4, 6), (6, 8) where 0 represents the input layer. The resulting performance was 0.79351 - a major increase from the previous more shallow model. In the future, I will refer to this model as MLP A.

My final multilayer perceptron model involved a few additional concepts. I will refer to this model as MLP E later in this paper. The overall structure consisted of 1 input layer, 3 hidden layers, and 1 output layer with sigmoid activation. After each of the hidden layers with ReLu activation was a dropout layer [12] with 50% probability of keeping the neuron. While initializing the model, another version of the model input was created by multiplying the model input by a set of randomly initialized weights from a normal distribution with a standard deviation of 0.01. The original model input was still fed into the input layer, but the modified input was added to the output of the second and the third hidden layer. Each of the hidden layers consisted of 4096 hidden neurons. The performance of this model came out to be 0.80180.

Based on the analysis on multilayer perceptron models' performances, it seems that the sigmoid activation function for the output layer performs significantly better than the softmax activation function. A possible reason may be because softmax ensures that the sum of the probabilities of the outputs is 1, whereas the sum of probabilities with sigmoid activation may exceed that and, therefore, contain additional labels that could result in a higher score. Based on the results, it also seemed that 2000-3000 hidden neurons, by themselves, resulted in some overfitting. Adding dropout layers after each of those layers tended to make the performance better. Lastly, adding residual connections seemed to drastically improve the performance of the neural networks. Possible experiments in the future may include testing more variations of the number of hidden neurons and layers, as well as adding more extensive and conclusion experiments on the difference between sigmoid and softmax activation functions.

### 3.4 Autoencoder

An additional model I used for experimentation purposes was the autoencoder neural network [11]. The reasoning behind such a choice was due to the fact that autoencoders are forced to learn a compressed representation of the input due to the number of hidden neurons being much less than the input and output size [11]. In my own implementation of the neural network, there is 1 input and output layer, and 2 hidden layers. The hidden layers contain 1152 and 300 hidden neurons, respectively. The idea is that the hidden layer containing the 300 hidden neurons would be forced to detect any significant structures or patterns in the data before sending them to the output layer. After just the training stage, the autoencoder model scored a GAP of 0.69. Due to time and resource constraints, the model was not continued with any further. Possible future experiments would be to complete the evaluation and inference processes, reduce the number of hidden layers to 1, and to modify the number of hidden neurons based on performance.

### 3.5 Convolutional neural network

Due to the extreme success in applying convolutional neural networks to image recognition problems, I decided to test their performance on the YouTube-8M dataset [9]. Since image data is essentially a 2-dimensional matrix with values inside, it's possible to manipulate the model input to be the same shape. For my initial implementation of a simple convolutional neural network, I first reshaped the model input to batch size x number of features x 1, where batch size corresponds to the hyperparameter, number of features corresponds to 1152 (both video and audio data), and 1 corresponds to the last dimension to fit into a convolution layer.

The simple structure of my convolutional neural network consists of a 2-D convolution layer with a kernel size of 1 and an output size of 32, a max-pooling layer with a kernel size of 1, a flattening layer, a fully connected layer with ReLu activation and 6000 hidden neurons, a dropout layer with 0.5 keep probability, and then an output layer with softmax activation. The final GAP score resulted in being 0.69569 - significantly less than the multilayer perceptron models.

Despite the poor performance of the convolutional neural network I implemented, I strongly believe improvements can be made to it. Possible improvements may be further manipulating the model input to be able to use a 3-D convolution layer, using larger kernel sizes to retain important information, increasing the output size of the convolution layer so that more of the structure of the data may be kept, and also increasing the depth of the network.

## 4. Ensemble approaches

To achieve a higher GAP score towards the end of the competition, I employed various ensemble approaches to my existing models. All of the ensemble methods that I used can be divided into 2 primary categories: ensemble learning with the models themselves or ensemble learning with the submission comma-separated values (CSV) files.

### 4.1 Ensemble learning with models

Having already experimented with many different types of models, I decided to see if combining them together might further improve the performance. Due to the submission CSV files containing the predicted labels and probabilities, I decided on an averaging method for the ensembles as opposed to others such as majority voting.

The first ensemble I created consisted of 4 neural networks. Each of the neural networks were the same: an input layer, 1 hidden layer with 2048 hidden neurons and ReLu activation, a dropout layer with 0.5 keep probability, and an output layer with sigmoid activation. The ensemble model trained all 4 of those neural networks in parallel, took the output predictions, summed them up, and then multiplied that total by 0.25 - essentially averaging the outputs. When training the neural networks individually, they each received a training GAP of approximately 0.71. Upon averaging the outputs, the training GAP resulted in being 0.74 - a slight increase.

The second ensemble approach also used the same base learning algorithm of the previous method, but instead it contained a stacking methodology. Instead of simply averaging the outputs of the 4 neural networks, I concatenated the outputs from the 4 models and fed them into another neural network as model input. The latter neural network consisted of an input layer, a hidden layer with 2048 hidden neurons and ReLu activation, a dropout layer with 0.5 keep probability, and then an output layer with sigmoid activation. The resultant training GAP score came out to be 0.65, which is substantially lower than both the previous ensemble approach and the single neural network.

Having seen that model averaging did boost the overall performance, I decided to test the concept on my best single model, the MLP E (as mentioned above). To experiment with various approaches, I built 2 ensembles - one consisting of 2 MLP E models and the other of 4 MLP E models. I also did modify the base learning rate to be 5E-4, just as I did with the base MLP E model. The final GAP score of the 4-model ensemble was 0.76885 and the final GAP score for the 2-model ensemble was 0.79143. The results of those 2 ensemble approaches proved to be surprising, especially since the initial 2 ensemble approaches showed promising results when averaging the outputs of models together.

To test the ensemble approach even more, I decided to instead use it on the MOE C model as mentioned previously. I also created 2 separate ensembles for the MOE C model - one with 5 models and another with 2 models. The final GAP score of the 5-model ensemble came out to be 0.77727 while the 2-model ensemble resulted in a 0.77686 final GAP score. Those 2 ensemble models with MOE C actually did show a performance improvement over the single MOE C model - although a minor increase.

Having seen the results of the model ensemble approaches, it's hard to draw any significant conclusions from such fusion approaches. Possible future experiments would be to test additional counts of the base model, possibly combine different models so that it's heterogeneous, and also to test other base learning algorithms in the case that there are specific ones that tend to benefit more from the ensemble approach.

### 4.2 Ensemble learning with submission files

In addition to ensemble approaches with the actual models, I also experimented with combining the submission CSV files generated from the model outputs.

The first attempt at this approach was with the submission files from the MLP E model and the Mixture of Experts model with 7 mixtures. As a reminder, the MLP E model submission file scored a final GAP of 0.8018 and the Mixture of Experts model with 7 mixtures submission file scored a final GAP of 0.79244. After averaging the outputs of both submission files, I received a CSV file that resulted in a 0.81133 final GAP score - an increase of .00953 from the highest scoring file.

To test the concept with additional models, I constructed an averaged submission file from the MLP E model, the Mixture of Experts model with 7 mixtures, and also the best performing logistic regression model - which scored a GAP of 0.75. After submitting the averaged file, I received a final GAP score of 0.80618. Despite the significantly weaker performance of the logistic regression model, the overall performance of the final submission file did not decrease that significantly; in fact, it still increased from the highest scoring model - MLP E.

Due to time and resource constraints, I was only able to experiment with one more. The final test combined the submission files from the MLP E model, the Mixture of Experts model with 7 mixtures, and also the MLP A model (as previously mentioned). For reference, each of the models had the following GAP scores:

MLP A: 0.80118
MLP E: 0.79244
Mixture of Experts with 7 mixtures: 0.79351

Upon averaging the 3 submission files together, the final CSV file received a score of 0.81587 - a .01469 increase from the MLP A model.

Possible additional experiments would involve testing many more different combinations of the models' submission files. There is strong evidence, from past research, that the less correlation there is among the models in the ensemble, the greater the increase in accuracy of the overall model [13]. One combination I believe would have yielded a much greater score increase, due to the diversity, is the MLP A model, MLP E model, Mixture of Experts model with 7 mixtures, and also a frame-level model such as the LSTM model provided.

## 5. Applications

By participating in the competition and sharing my results, I am extremely grateful to have been given the opportunity to participate in such ground-breaking research. I am excited for the future applications that advanced video recognition is able to bring about. Based on my own research, I have found the following areas to be ripe subjects capable of being revolutionized by scientific advances in video understanding: video recommendation and search, safety and security, transportation, robotics, and video analytics.

### 5.1 Video recommendation and search

A prime example of a video recommendation situation is YouTube. YouTube poses one of the most complex and daunting video recommendation problems to the scientific community due to its sheer size and scale. It's estimated that, since its launch in 2005, YouTube contains over 45,000,000 videos - a number that is constantly growing [2]. Every minute, an estimated seven hours of video is being uploaded to the massively popular site for everyone [2].

According to the current publications on the YouTube recommendation algorithm, graph structures were extensively used in the user recommendations until deep learning approaches were discovered to perform better on such a problem [2, 3]. Based on [3], two neural networks are effectively used in the process - one for candidate generation and another for ranking. The candidate generation network essentially generates a list of possible suggestions to the user. The generation network then sorts through those recommendations to assign specific rankings to the videos based on the user's history and preferences. The inputs to the networks are typically the video IDs, search query tokens, viewer demographics, and co-viewer statistics.

A notable observation I noted from reading through the papers is that the recommendation algorithms do not directly suggest videos based on the actual content of the video. Instead, much of the suggestions are based on the statistical summaries computed from co-viewing habits and personal viewing history. Immediately I am able to see the potential for video understanding research from the YouTube-8M dataset. With increased progress in building models to understand the actual content and subject matter of videos, I strongly believe video recommendation can be made significantly better than it already is. As a YouTube user myself, there are still times when I am confused as to why I am recommended a specific video that seems to be completely unrelated to the ones I've been watching. Rather than solely depending on video viewing statistics, demographics, video IDs, search query tokens, and co-viewer statistics, adding inputs based on the actual content of the videos will significantly boost the relevance of suggested videos. As the technology gets even more advanced at recognizing specific objects and actions within each video, even more detailed analytics may be garnered and thus leading to the optimal recommendations.

Video search is another strongly related problem that is entirely capable of being solved with advanced video understanding techniques. If the models were able to understand the specific content in each video and generate advanced analytics on each one, traditional search algorithms would be able to parse such statistics to search for related objects.

### 5.2 Safety and security

Another area ripe for improvement is video surveillance. With the increasing mobility and advances in technology today, increased security is also a necessity for many to feel comfortable. Even today, there are still human security personnel manually watching security camera footage for any suspicious behavior. It's a well-known fact that human attention to detail is substantially decreased as time progresses, which then also reduces the chances of a human detecting unusual circumstances or responding to immediate threats [14].

Much of the security cameras today are used for two purposes: real-time threat detection and forensic investigation. For both purposes, identity tracking, location tracking, and activity tracking are 3 important features to keep note of in video surveillance [14]. Identity tracking helps to see who the person or what the vehicle is. Location tracking helps to see where the event is occurring, and activity tracking detects what exactly is happening. Despite the many advances in saliency detection and camera technology, much can still be improved for all 3 features with increased research in video understanding [14].

Activity tracking is one feature that is extremely closely related to the content of the YouTube-8M competition. Being able to detect the specific action in videos is crucial for understanding the world in front of the camera. Location tracking and identity tracking are also closely to image recognition, but may also be significantly improved with the dimension of time added into video. With time added, additional analytics are able to be gained such as the pace at which the subject is travelling and the rate of activity occurring in the surrounding environment.

### 5.3 Transportation

Self-driving cars are all the rage right now as various companies all vie for position in the technology that could disrupt all of transportation. As of now, there are various different ways that companies are using to essentially allow the vehicle to "see" the world around them - LIDAR (Light detection and ranging), radar, and vision techniques [15].

The choice of using video techniques has proven to be the cheaper alternative - with only a few cameras necessary for the car to view the world around them [15]. Despite such advances, the object and content detection occurring in such technologies are typically using image recognition models. With the addition of video recognition models, such vehicles are then able to obtain abilities akin to a human eye and be able to see their surroundings in real-time. With such an increase in visual senses, more data is able to be aggregated to eventually result in safer and more reliable self-driving vehicles.

### 5.4 Robotics

Similar to self-driving vehicles, robots may also greatly benefit from abilities akin to having a human eye. According to [1], the researchers were trying to teach robots to learn manipulation actions from simply watching videos - which could potentially lead to extraordinary results. Based on [1], two different convolutional neural networks were used: one was used to classify the hand grasp type, while the other was used for object detection [1]. The researchers only categorized the hand grasp types into six different categories, depending on the situation. Imagine a human eye only being able to view the world in step-by-step images. To learn to grab an item, a human would then only have a limited number of specific grasp types to pick and only a limited way to hold an item. What about the intricate details that go into hand movements and real human interactions?

Simply put, images are only capable of providing blocks of information with no notion of time sequence or connection. With videos, data about each specific detail of a movement and of the exact timing of human interactions is able to be captured within the sequence of frames. With advancements in video understanding, we will eventually be able to reach that point. In order for robots to perform human chores such as doing the laundry, cleaning surfaces, and cooking meals, they need to be able to view their surroundings as human do in a continuous and sequential manner.

### 5.5 Video analytics

We live in a world today where data is the key to truly revolutionary technological advancements. Perhaps the most important application of video understanding research is being able to supply data for everything and anything ever recorded through video.

Having the ability to understand the content of videos, technology will be able to assist us with tools such as automatic generation of "table of contents" for videos and video descriptions. By being able to analyze the entire video and distinguish between the separate pieces, it will be possible to have a section list automatically generated for us so that it's possible to immediately skip to a specific time in the video, rather than having to manually search for it. Once it's possible to analyze each of the smaller sections of the video, models will be able to generate specific descriptions for the one activity/event occurring in that section. Then with a high-level model, all of the individual video section's descriptions could be parsed to generate an overall description for the entire video automatically.

## 6. Future work

Video recognition is an exciting area of research that has the potential to improve many of the existing technologies that we have today. Due to my lack of time and resources, I was not able to experiment or research all that I could have.

A couple public models that I would have liked to experiment with are the VGG16 [16] and GoogLeNet [17]. Each of those models have proven their exemplary image recognition accuracy in the past, but I am very curious as to how their performance on the YouTube-8M dataset may be.

An additional idea I have been interested in exploring is extracting other features from YouTube videos to see if they may increase the power of existing models. Based on my reasoning, I believe it may be best to extract features that are universal to all video formats so that the data wouldn't solely be limited to YouTube. Some possible areas for extraction may be the video length, the level of activity in the video (perhaps measured by RGB fluctuations), and the date of video publication.

## 7. Conclusion

In this paper, I provided an overview of the YouTube-8M Video Understanding Challenge hosted as a Kaggle competition in the first two sections. Then, I discussed my approach to the competition and provided the specific performance scores for each of my models. Towards the end, I also detailed each of the model ensemble approaches that I did to provide a significant boost in my overall competition score. Lastly, I discussed several possible applications of the YouTube-8M dataset and my ideas. Video understanding is a very exciting area of research right now, and I am extremely grateful to have been able to participate in such a competition meant to push the boundaries in the subject matter.

I would like to thank Google for providing the YouTube-8M Tensorflow Starter Code, which has helped me tremendously as this was my first jump into the fascinating world of machine learning.